\documentclass[10pt,twocolumn,letterpaper]{article}

\usepackage{cvpr}
\usepackage{times}
\usepackage{epsfig}
\usepackage{graphicx}
\usepackage{amsmath}
\usepackage{amssymb}
\usepackage{multirow}

\usepackage[pagebackref=true,breaklinks=true,letterpaper=true,colorlinks,bookmarks=false]{hyperref}

 \cvprfinalcopy 

\def\cvprPaperID{****} 

\ifcvprfinal\fi
\begin{document}

\title{Creating Artificial Modalities to Solve RGB Liveness
}

\author{Aleksandr Parkin\\
VisionLabs\\
{\tt\small a.parkin@visionlabs.ai}
\and
Oleg Grinchuk\\
VisionLabs\\
{\tt\small o.grinchuk@visionlabs.ai}
}

\maketitle
\thispagestyle{empty}

\begin{abstract}
Special cameras that provide useful features for face anti-spoofing are desirable, but not always an option. In this work we propose a method to utilize the difference in dynamic appearance between bona fide and spoof samples by creating artificial modalities from RGB videos. We introduce two types of artificial transforms: rank pooling and optical flow, combined in end-to-end pipeline for spoof detection. We demonstrate that using intermediate representations that contain less identity and fine-grained features increase model robustness to unseen attacks as well as to unseen ethnicities. The proposed method achieves state-of-the-art on the largest cross-ethnicity face anti-spoofing dataset CASIA-SURF CeFA (RGB).
\end{abstract}

\section{Introduction}
Recently, face anti-spoofing research received great attention due to rapidly growing integration of face recognition systems \cite{phillips2018face} to human lives (biometric payments, physical access control systems, etc.). Given importance of these systems, providing high level security from fraudulent attacks is a must.

One way to solve face anti-spoofing task is to support RGB stream with additional modalities (infrared, depth, thermal images). These modalities provide useful information for liveness detection allowing to create successful systems \cite{Parkin2019RecognizingMF} on relatively small amounts of data. However, there are plenty of cases where one cannot use extra sensors due to cost related or other reasons. For example, establishments with already placed RGB cameras are unlikely to change all hardware in favor of new multi-modal protection method. Therefore, RGB-only face anti-spoofing problem remains a hot topic nowadays.

The main issue with liveness detection is the difference between training and testing use cases. It is nearly impossible to examine a system in all potential scenarios. Intruders can create new unseen artefacts or bona fide persons may have appearance different from the one presented in the training set. This is especially severe in the case of RGB data, methods that utilize raw RGB images tend to overfit to the selected set, showing insufficient performance in unseen scenarios ~\cite{liv04,liv05,liv06,liv07}.

\begin{figure}
\begin{center}
   \includegraphics[width=0.9\linewidth]{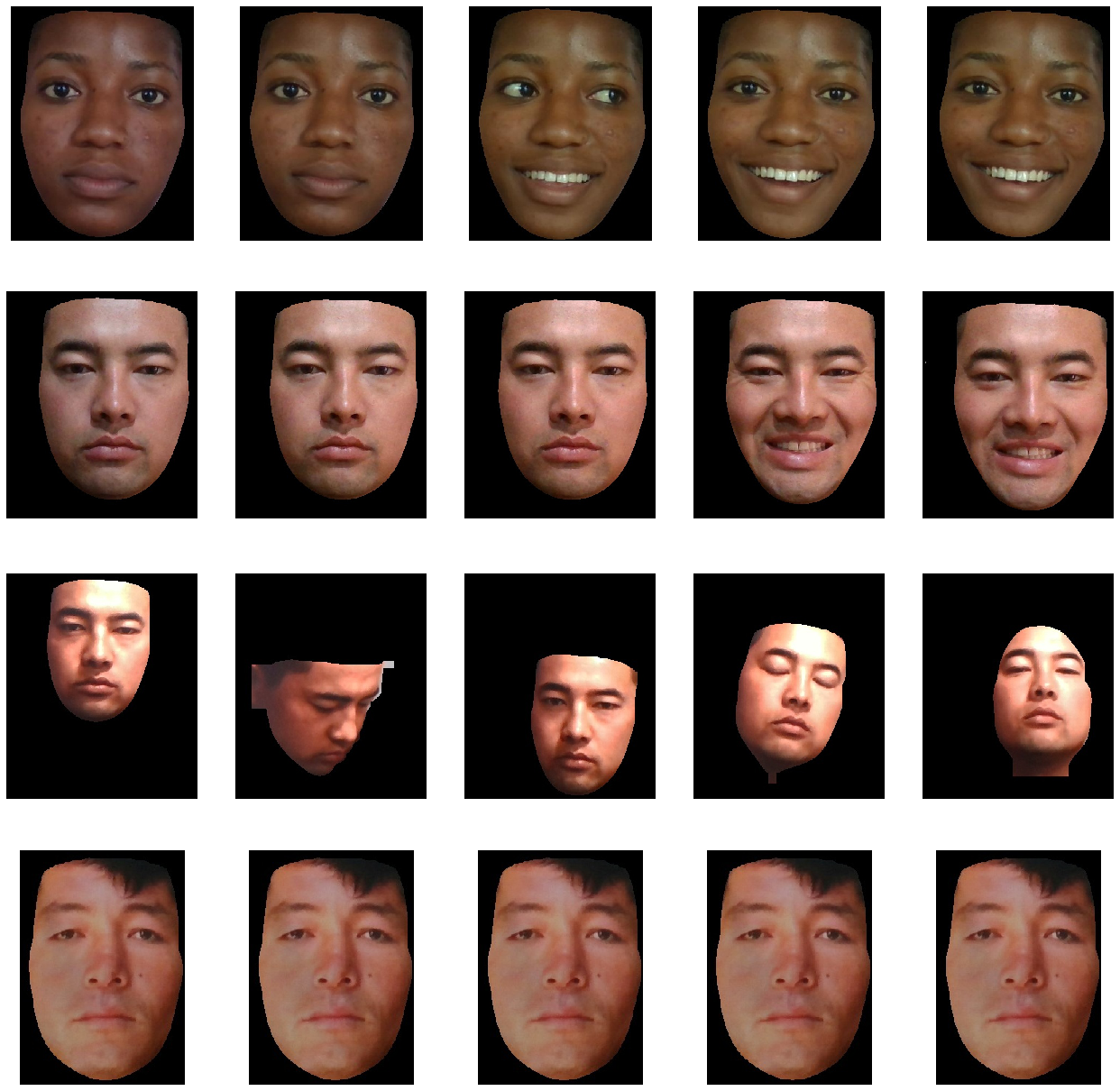}
\end{center}
   \caption{Samples from CASIA-SURF CeFA dataset. Each row indicates 5 uniformly distributed images from one RGB track. First row - bona fide (train), second row - bona fide (test), third row - replay attack (train), fourth row - printed attack (test).}
\label{fig:fake_types}
\end{figure}

Another barrier towards reliable face anti-spoofing system is the limited number of subjects in the liveness datasets. Collecting anti-spoofing samples could only be achieved in laboratory scenario while for face recognition there's tons of images available online. And until recently, lack of subject count and diversity prevented from using deep learning approaches on raw RGB frames as well as from utilizing temporal features from video sequence.

CASIA-SURF CeFa dataset \cite{CASIA_CeFA} is the largest available face anti-spoofing dataset in terms of subjects, ethnicities and fake types. Introduced testing protocols measure the performance in cross-ethnicity and cross-attack settings, addressing the problem of unseen scenarios. Moreover, number of presented subjects allow to move problem statement from the frame level to the video level.

Distinguishing between videos of bona fide and spoof examples is easier since we can utilize the benefits from face appearance changes in time. Natural behavior of real subjects in front of camera implies micro face movements which differ from attack presentations. Treating the whole video as a one sample allows us to create new representations of the sample and store useful information in it. Ultimately, these representations are similar to the images from special cameras. For example, disparity map from stereo camera provides a 3D surface which is extremely useful in identifying flat spoofing artifacts such as printed photos. On the other hand, video of a real subject moving his head could be transformed to a depth-like image through applying Optical flow \cite{OFLiu} to two diverse frames from that video. Here, optical flow image could be treated as a new artificial modality that provides useful features for liveness task.

Despite the fact that CASIA-SURF CeFa dataset allows to consider the problem at a video level, the amount of data is still small for naive raw rgb approaches. In this paper we reduce the problem of video sequence classification to a problem of image classification by introducing artificial modalities with rich features, such as optical flow and rank pooling \cite{RankPooling}. We also propose a sequence augmentation that trasnforms bona fide track to a fake track, enlarging the number of spoof samples. Finally, we use a very simple fusion neural network architecture with shallow backbones to combine different modalities together.

As a result, our method achieves state-of-the-art on the largest cross-ethnicity face anti-spoofing dataset CASIA-SURF CeFa (RGB), taking first place in \href{https://sites.google.com/qq.com/face-anti-spoofing/winners-results/challengecvpr2020}{Chalearn Single-modal Face Anti-spoofing Attack Detection Challenge at CVPR 2020} and overcoming second place by a factor of $1.8x$.

\begin{figure}
\centering
\includegraphics[width=0.99\linewidth]{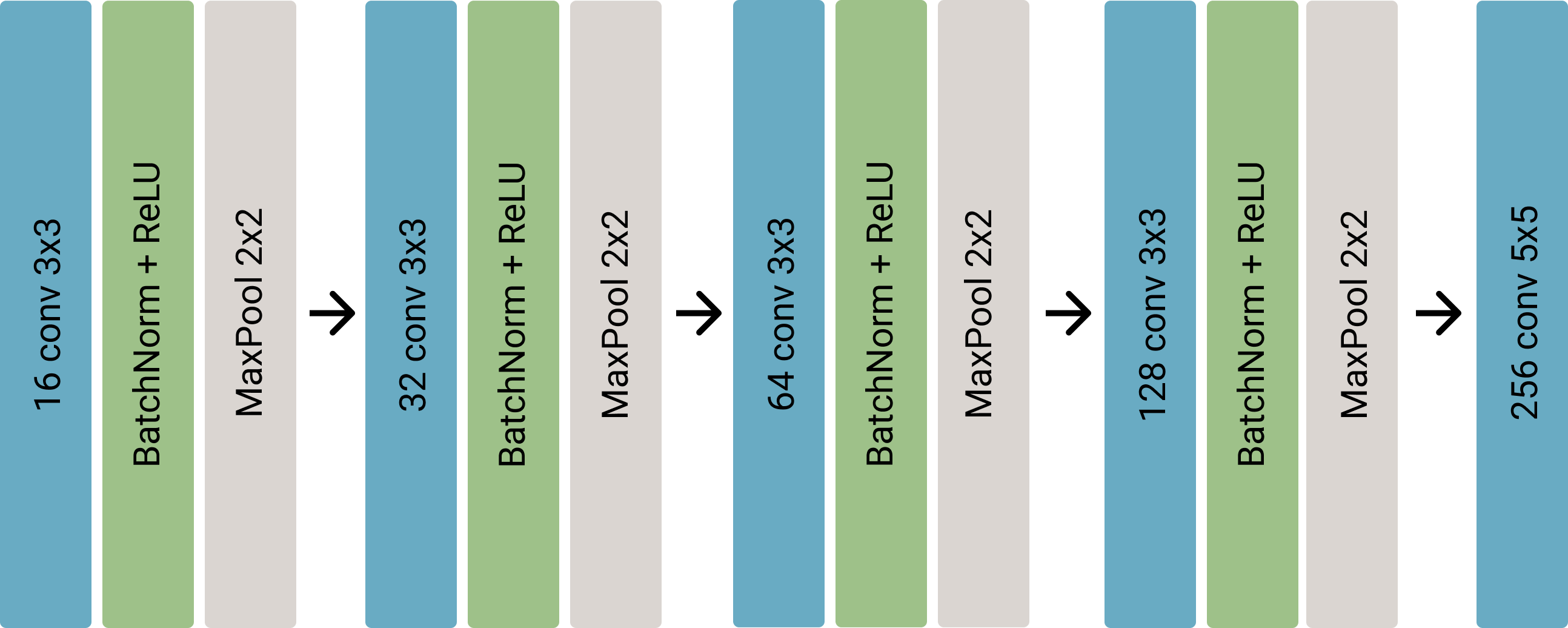}
\caption{SimpleNet architecture: 4 blocks of Conv $3\times3$- BatchNorm - Relu - MaxPool of sizes 16, 32, 64, 128 followed by Conv $5\times5$ with 256 filters.}
\label{fig:simplenet}
\end{figure}

\section{CASIA-SURF CeFa dataset}
CASIA-SURF CeFa dataset includes 1607 subjects, 3 ethnicities and 4 different attack types, including 3D masks. For the Chalearn Single-modal face anti-spoofing challenge organizers introduced 3 protocols, where training and testing sets contained diverse ethnicities and attack types. For each protocol training set included 200 real and 200 fake videos and testing set included 400 real and 1800 fake videos. The videos were presented in a form of frame sequences with removed background and aligned faces. Fig. \ref{fig:fake_types} shows some examples from the dataset. Both bona fide and spoof images look very different for train and test part, enforcing challenge participants to use methods that can be generalized well to unseen samples.

\textbf{Evaluation metrics.}
To measure the performance on the test set, the Average Classification Error Rate (ACER) is used. It includes Attack Presentation Classification Error Rate (APCER) and Bona Fide Presentation Classification Error Rate (BPCER). ACER can be calculated as $ACER = (APCER+BPCER)/2$, where $APCER = FP / (TN + FP)$ and $BPCER = FN / (TP + FN)$. Here $TP$,$TN$,$FP$,$FN$ are number of True Positive, True Negative, False Positive and False Negative samples respectively. Decision threshold is calculated on a development set.

\section{Proposed Method}
This section describes proposed method. The same pipleline is applied to all three protocols so for the sake of simplicity we further consider only one protocol. We denote a sequence of RGB frames with face as a \textit{track} and an image obtained from the track using some transformation as
\textit{artificial modality}.

\begin{figure}
\centering
\includegraphics[width=0.99\linewidth]{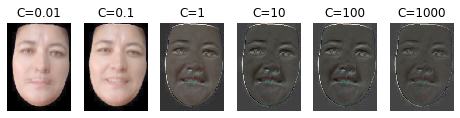}
\caption{Rank pooling transform with different level of regularization parameter $C$.}
\label{fig:rankpool}
\end{figure}

\subsection{Artificial modalities}
Due to huge difference in training and testing images (both fake and real), we decided not to train on raw RGB images and substitute them with artificial modalities. Good artificial modalitiy should contain fewer fine-grained image details (to prevent from overfitting) but at the same time include some additional useful information for liveness detection task. Namely, we have chosen Optical Flow and RankPooling which both fulfil the requirements mentioned above.

\begin{figure*}
\centering
\includegraphics[width=0.99\linewidth]{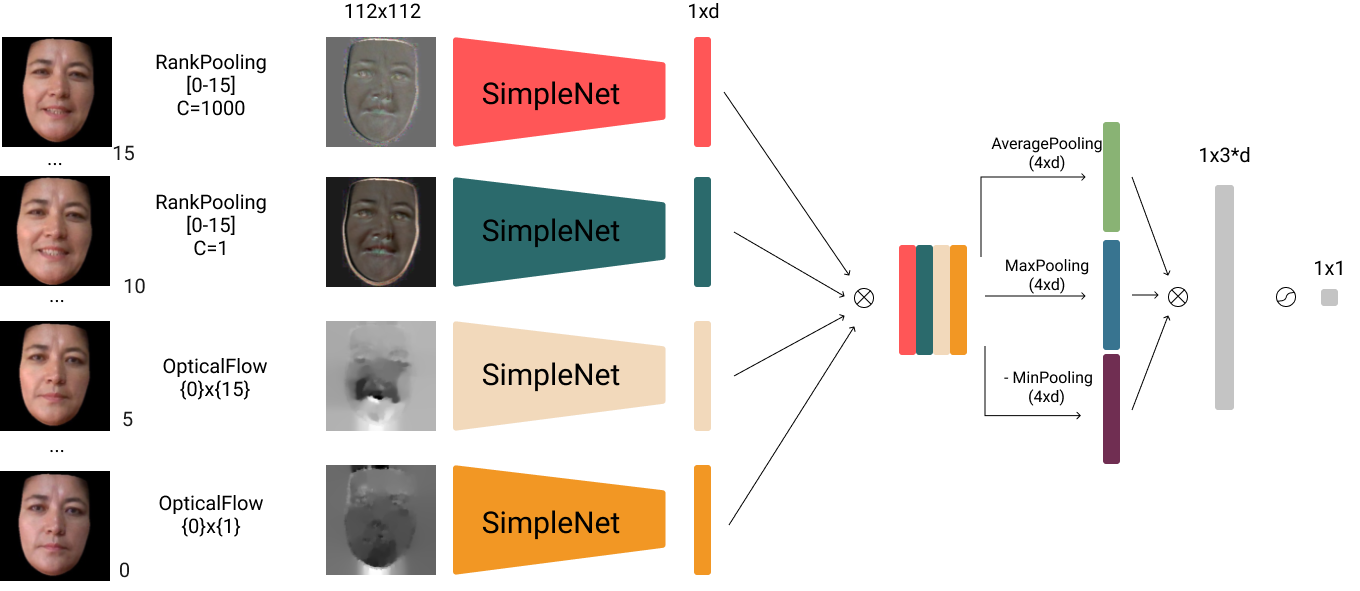}
\caption{Method diagram. 16 uniformly selected images from track are used to extract 4 "modalities": 2 RankPooling and 2 Optical flows with different params. Then these modalities are processed by different instances of SimpleNet followed by min, max, avg poolings and one fully connected layer with sigmoid nonlinearity.}
\label{fig:architecture}
\end{figure*}

\textbf{Optical flow modality} is based on a non deep learning algorithm proposed in \cite{OFLiu}. We obtain an optical flow from two pairs, one between first and last images from the track, the other is between first and second track images. The motivation behind is to show the network that flow for real track will change if we select images at a different time,  while flow for fake track should remain approximately the same. Moreover, flow for bona fide track could show the rough 3D structure of movable face parts such as eyes or mouth.

\textbf{Rank pooling modality} encodes video sequence into a feature vector through optimization process which can be formulated as a Support Vector Regression task \cite{RankPooling}. It creates a dynamic/temporal image from the whole RGB track, capturing the evolution of frame-level features over time.

Given hyperparameter C, we chose two different values: $C=1$ and $C=1000$, which imply high and low levels of regularization constraints in SVR, therefore leading to different solutions. The corresponding rank pooling images looks differently (Fig. \ref{fig:rankpool}): $C=1$ preserves more identity information while $C=1000$ contains face shape changing along the images.

\subsection{Sequence Augmentation}
Due to huge difference between replay and printed fakes (Fig. \ref{fig:fake_types}) we introduce a simple yet effective sequence transform that is aimed at creating more diversity in training set. Given a track, we create a new one by replicating a random image  and changing track label to "fake". This creates a new family of fake tracks that look more similar to the printed fakes than the replay data. The donors for sequence augmentation are selected both from real and fake original tracks.

\subsection{Architecture}
To further classify extracted artificial modality images, we introduce a fusion architecture with SimpleNet (Fig. \ref{fig:simplenet}) backbones. The proposed backbone is deep enough to process images with removed identity information but in the same time shallow enough to avoid overfitting.

Each of 4 obtained tensors (2 from optical flow modality and 2 from rank pooling modality) are processed by separate SimpleNet backbones that return embeddings of size $d=256$. Embeddings are concatenated to the tensor of shape $4\times d$. After that Max, Avg and Min pooling are applied among the first dimension and final feature vectors are concatenated, getting $3 \times d$ tensor. We flatten this tensor, process it with fully connected layer followed by sigmoid nonlinearity and use standard Binary Cross Entropy loss for training.

On a contrary to simple concatenation, proposed fusion method (with Max, Avg and Ming pooling) keeps the same number of input features to fully connected layer (always $3 \times d$), making the framework scalable to different number of input modalities. Moreover, it is also better than just summation (which is equivalent to avg pooling), since features from different modalities may vary in importance given different inputs. 

\begin{table*}
\begin{center}
\begin{tabular}{|c|c|c|c|}
\hline
Method & APCER, $\%$ & BPCER, $\%$ & ACER, $\%$\\
\hline\hline
Baseline & 21.83 $\pm$ 1.70 & 25.20 $\pm$ 22.00 & 23.42 $\pm$ 12.14 \\
RankPool(C=1000) & 14.11 $\pm$ 13.52 & 11.25 $\pm$ 12.75 & 12.68 $\pm$ 4.39 \\
+Sequence Augmentation & 0.68 $\pm$ 0.21 & 13.91 $\pm$ 10.03 & 7.30 $\pm$ 5.00 \\
+RankPool(C=1) & 1.07 $\pm$ 0.53 & 13.00 $\pm$ 10.75 & 7.03 $\pm$ 5.20 \\
+OpticalFlow & 0.11 $\pm$ 0.11 & 5.33 $\pm$ 2.37 & 2.72 $\pm$ 1.21 \\
\hline
\end{tabular}
\end{center}
\caption{Results on CASIA-SURF CeFa test subset.}
\label{tab:ablation}
\end{table*}

\section{Experiments}
In this section we describe our experiments on CASIA-SURF CeFa dataset and show individual effect of proposed contributions.

\subsection{Implementation details}
Code is available at \href{https://github.com/AlexanderParkin/CASIA-SURF_CeFA}{github}.
All code is implemented in pytorch and run on a single NVIDIA RTX 2080 GPU. We train model with batchsize 32 and number of threads 8. Training process requires 1.5G of GPU memory. We train a model for 5 epochs (each epoch contains 20x training data) via Adam algorithm with learning rate of 0.0001. Total training time is 1h for one protocol, we use the same pipeline for all 3 protocols.

\subsection{Pipeline}
Denote images from track $k$ as $\{X_i^k\}$, where $i=0 .. len(k)-1$ and $t^k=\{0,1\}$ - label of track, where $0$ - fake, $1$ - real. Then for each track we select $L=16$ images  uniformly, obtaining $\{X_j^k\}$, where $j=0 .. 15$ (For example, if track contains 48 images, we will select every third image).

Then with probability 0.5 we apply Sequence Augmentation (SA), e.g. select one random image from $\{X_j^k\}$ and replace all 16 images in track with it, also setting  $t^k$ to 0.  After that we remove black borders and pad image to be square of size $112\times 112$. Then we apply intensive equal color jitter to all images, emulating different skin color.  

Given $\{X_j^k\}$, we apply rotation, color jitter and random shift to each image independently. Then we apply 4 modality transforms: RankPooling ($\{X_j^k\}$, C=1000), RankPooling ($\{X_j^k\}$, C=1), Flow($X_0^k$, $X_{15}^k$), Flow ($X_0^k$, $X_1^k$). These transforms return 4 tensors with sizes $3\times 112\times112$, $3\times 112\times112$, $2\times 112\times112$, $2\times 112\times112$ respectively.  See Figure~\ref{fig:architecture} for details.

In the following experiments we assume that this pipeline remains the same except explicitly mentioned changes.

\subsection{CASIA-SURF CeFa}
\textbf{Baseline.}
To set up basic performance, we first train only on raw RGB image pairs and without Sequence Augmentation. We concatenate first and last images from track, obtaining tensor of size $6\times112\times112$. This can also be treated as an artificial modality and all further processing is the same as in final method. While such modality can force the network to learn some kind of optical flow inside it, this method overfits and reaches $23.42\%$ ACER on a testing set (Tab. \ref{tab:ablation}). Huge standard deviation among BPCER protocols scores indicates instability of this modality on different ethnicities.

\textbf{Rank pooling.} To show the efficacy of carefully selected artificial modalities, we substitute the naive baseline with RankPooling (C=1000). The test set error drops to $12.68\%$ proving that dynamic features without fine-grained information are better than raw RGB data.

\textbf{Sequence augmentation.} This experiment demonstrates that simple data augmentation can play a vital role in model performance. Adding SA (Tab. \ref{tab:ablation}) reduces ACER rate to $7.3\%$ comparing to the previous experiment. However, using this transform alone won't help. Without small inter-track perturbations (color jitter, rotation, shift) modality will collapse to a still image. 

Adding one more RankPooling modality with C=1 yields to small increase of final score ($ACER=7.03\%$) so we decided to stop with only two representatives from rank pooling family.

\textbf{Optical flow.} Adding optical flow to the pipeline led to the state-of-the-art result on CASIA-SURF CeFa RGB dataset - $2.72\%$ ACER. The optical flow images emphasize the difference in face mimic movements between bona fide and spoof tracks. Static presentation attacks are more calm if we look at them in optical flow "spectrum". It is proven by $APCER=0.11\%$, e.g. 2 out of 1800 fake samples are misclassified. Bona fide errors are higher: $BPCER=5.33\%$, where most errors go to real tracks with negligible face movements. These scores show that artificial modalities have a drawback - when we selected optical flow and rank pooling, we implied that bona fide subject will change his mimics through time, which is not always true. 

\section{Conclusion}
In this paper, we proposed a method to solving face anti-spoofing through creating artificial modalities and sequence augmentation. We showed that careful selection of intermediate data representations, such as rank pooling  with different regularization parameters or non trainable optical flow decrease the risk of overfitting and improve performance compared to the raw rgb approaches. We also introduced an effective network architecture that is capable of fusing arbitrary number of input modalities. Finally, we demonstrated a simple trick to enrich the collection of fake tracks. As a result, our model achieved first place in Chalearn Single-modal Face Anti-spoofing Attack Detection Challenge at CVPR 2020.

{\small
\bibliographystyle{ieee_fullname}
\bibliography{egpaper_final}
}

\end{document}